%% file: main.tex
\documentclass[conference]{IEEEtran}
\IEEEoverridecommandlockouts
\usepackage{cite}
\usepackage{amsmath,amssymb,amsfonts}
\usepackage{algorithmic}
\usepackage{graphicx}
\usepackage{textcomp}
\usepackage{xcolor}
\usepackage{hyperref}
\def\BibTeX{{\rm B\kern-.05em{\sc i\kern-.025em b}\kern-.08em
    T\kern-.1667em\lower.7ex\hbox{E}\kern-.125emX}}
\begin{document}

\title{Intrinsic Numerical Robustness and Fault Tolerance in a Neuromorphic Algorithm for Scientific Computing \\

}

\author{\IEEEauthorblockN{Bradley H. Theilman }
\IEEEauthorblockA{\textit{Neural Exploration \& Research Laboratory} \\
\textit{Sandia National Laboratories}\\
Albuquerque, NM \\
bhtheil@sandia.gov}
\and
\IEEEauthorblockN{James B. Aimone}
\IEEEauthorblockA{\textit{Neural Exploration \& Research Laboratory} \\
\textit{Sandia National Laboratories}\\
Albuquerque, NM \\
jbaimon@sandia.gov}

}

\maketitle

\begin{abstract}

The potential for neuromorphic computing to provide intrinsically fault-tolerant has long been speculated, but the brain's robustness in neuromorphic applications has yet to be demonstrated.
Here, we show that a previously-described, natively spiking neuromorphic algorithm for solving partial differential equations is intrinsically tolerant to structural perturbations in the form of ablated neurons and dropped spikes. The tolerance band for these perturbations is large: we find that as many as 32\% of the neurons and up to 90\% of the spikes may be entirely dropped before a significant degradation in the accuracy results. Furthermore, this robustness is tunable through structural hyperparameters. This work demonstrates that the specific brain-like inspiration behind the algorithm contributes to a significant degree of robustness expected from brain-like neuromorphic algorithms. 
\end{abstract}


\section{Introduction}

Algorithm design generally assumes that the underlying computing hardware that implements an algorithm's deterministic operations does so reliably and accurately with a well-defined precision. In the real-world, computing hardware is physically realized and thus is susceptible to noise. While the impact of intrinsic (e.g., thermal) or extrinsic (e.g., cosmic rays) noise on modern digital electronics is exceedingly small due to error-correcting circuit design and advanced fabrication processes, the impact of noise on emerging computing technologies is a major consideration. For example, noise-tolerant algorithms \cite{bharti2022noisy} are increasingly important for today's quantum computing systems that are largely limited by noise \cite{proctor2022measuring}. In this context, the role of hardware faults and errors in spiking neuromorphic algorithms remains relatively underexplored. However, a one-size-fits-all approach to understanding spiking neuromorphic algorithm robustness is likely impossible due to differences in neuromorphic hardware (e.g., analog versus digital), variability in spiking algorithm design, and application demands. For this reason, it is likely necessary to explore algorithm robustness in a case-by-case manner.

One class of neuromorphic computing applications for which numerical robustness is critical is its potential for use in scientific computing applications \cite{aimone2022review}. One of the most important uses of computers is for predicting the behavior of physical systems, which are often described by partial differential equations (PDEs) that do not have analytic solutions and so must be solved numerically. The finite element method (FEM) is one of the most important numerical techniques for solving PDEs on computers. FEM is a mathematical process for discretizing the problem of interest, yielding a large, sparse linear system. To yield accurate approximations for PDE solutions, finite element discretizations must have high resolution. It is not uncommon for ``production" finite element problems to involve linear systems with hundreds of millions or even billions of variables. Indeed, solving these kinds of equations has been a primary driver of innovation in high-performance computing (HPC). This leads to extremely large, sparse linear systems which must be solved on large-scale parallel computers, or, potentially, on large neuromorphic systems.

Because of the high computational demand required by high-resolution finite element simulations, such simulations are currently restricted to large HPC clusters housed in controlled environments. However, it is conceivable that a number of applications would benefit from high-resolution finite element simulations ``at the edge". Current computational technology limits the practical size of such simulations in edge devices. Neuromorphic computing may provide a pathway for increasing the resolution of edge simulations, enabling new applications, but would have to contend at the edge with an increased likelihood of faults in computation or communication that would be fatal for existing algorithms. Thus, the robustness of neuromorphic algorithms is of pressing concern. 

Recent work \cite{theilman2025solving} demonstrated a natively-neuromorphic spiking algorithm for solving partial differential equations (PDEs) with the finite element method. Our spiking finite element algorithm has built-in redundancy in that individual variables are necessarily over-represented among neurons and spikes. Here, we show that this redundancy has clear advantages in terms of the robustness of the algorithm: we find that the algorithm can sustain up to a 32\% loss of neurons and up to a 90\% loss of spikes without significant accuracy loss. Furthermore, our algorithm's robustness is tunable via structural hyperparameters. This work shows that brain-inspired algorithms can confer a significant error tolerance for numerical computing.

\section{Fault-tolerant Computing and Neuromorphic Solutions}

For many scientific computing \cite{aimone2022review} and edge computing \cite{schuman2022opportunities} applications of neuromorphic it is important to understand how a computation suffers if the underlying computing hardware exhibits faults or errors. This is particularly important as neuromorphic systems begin to reach large scales \cite{kudithipudi2025neuromorphic}, in which even low-frequency device errors may manifest themselves as moderately likely events in the lifetime of a simulation. 

Errors and faults have long been an important consideration in large-scale computing platforms \cite{canal2020predictive, cappello2014toward}. While modern conventional processors and memories are extremely reliable; the extreme scale of data centers and large high-performance computing (HPC) systems makes it a non-zero probability that large jobs may expect at least one, if not several, bit errors \cite{schroeder2009dram}. The increased frequency of such errors requires not only the use of advanced memory using error correcting codes (ECC), but also application-side mitigation such as more frequent checkpointing of large simulations \cite{egwutuoha2013survey}, both of which add considerable costs to scalable HPC systems. 

For a number of reasons, systems deployed at the edge should be expected to have much greater risk of errors, both due to more extreme environmental conditions, less quality control compared to HPC systems, and more risk of electrical and magnetic interference. At the same time, the cost--benefit of error correction in edge systems is more pronounced: edge systems are often high-consequence, thus demanding reliability and confidence, while typically resource limited, making the added costs of error correcting codes more restrictive.

Because of the perception that the brain is intrinsically robust to noise, neuromorphic systems have been proposed as a solution for more efficient, noise-tolerant computing. However, much of the previous research in this area has focused on noise added by the neuromorphic platform itself. For instance, the use of analog devices within resistive memory crossbars introduces noise that complicates their use in vector--matrix multiplication operations \cite{bennett2019wafer}, and thus many crossbar-based neuromorphic designs are intended to mitigate the additional noise added by analog components \cite{bennett2020evaluating, liu2018design, wu2020fault}. Similarly, hardware implementations of spiking neural networks (SNNs) can introduce challenges as well \cite{yerima2023fault}. While digital spiking systems may be highly reliable at a computational logic level, asynchronous communication between cores and chips can introduce a non-determinism in spiking communication (particularly in TrueNorth \cite{akopyan2015truenorth} and SpiNNaker's \cite{davies2012population, hoppner2021spinnaker} routing strategies). The risks of this broader asynchrony at scale can be mitigated with strategies such as global barrier synchronization, as on the Loihi platform \cite{davies2012population}, though this too introduces costs in computation \cite{li2025deterministic}.

Arguably, common to these neuromorphic considerations around fault-tolerance betrays an intrinsic fallacy in the field: we engineer hardware inspired by the brain with properties that generally increase non-determinism in computation and communication, yet we offset the benefits of that hardware by working to correct the noise. Rather, why should the neuromorphic hardware community focus on developing more reliable components if the brain is highly energy-efficient in part because it can compute reliably in spite of unreliable components? Stated differently, in principle neural algorithms should be intrinsically tolerant to modest levels of noise without forcing hardware design choices that incur additional costs. 

Stated differently, one question facing the neuromorphic field is as follows:\textbf{\textit{ is it possible to compute reliably with unreliable components?}}

\section{Neuromorphic Finite Element algorithm}

\begin{figure}[htbp]
\centerline{\includegraphics[width=0.5\textwidth]{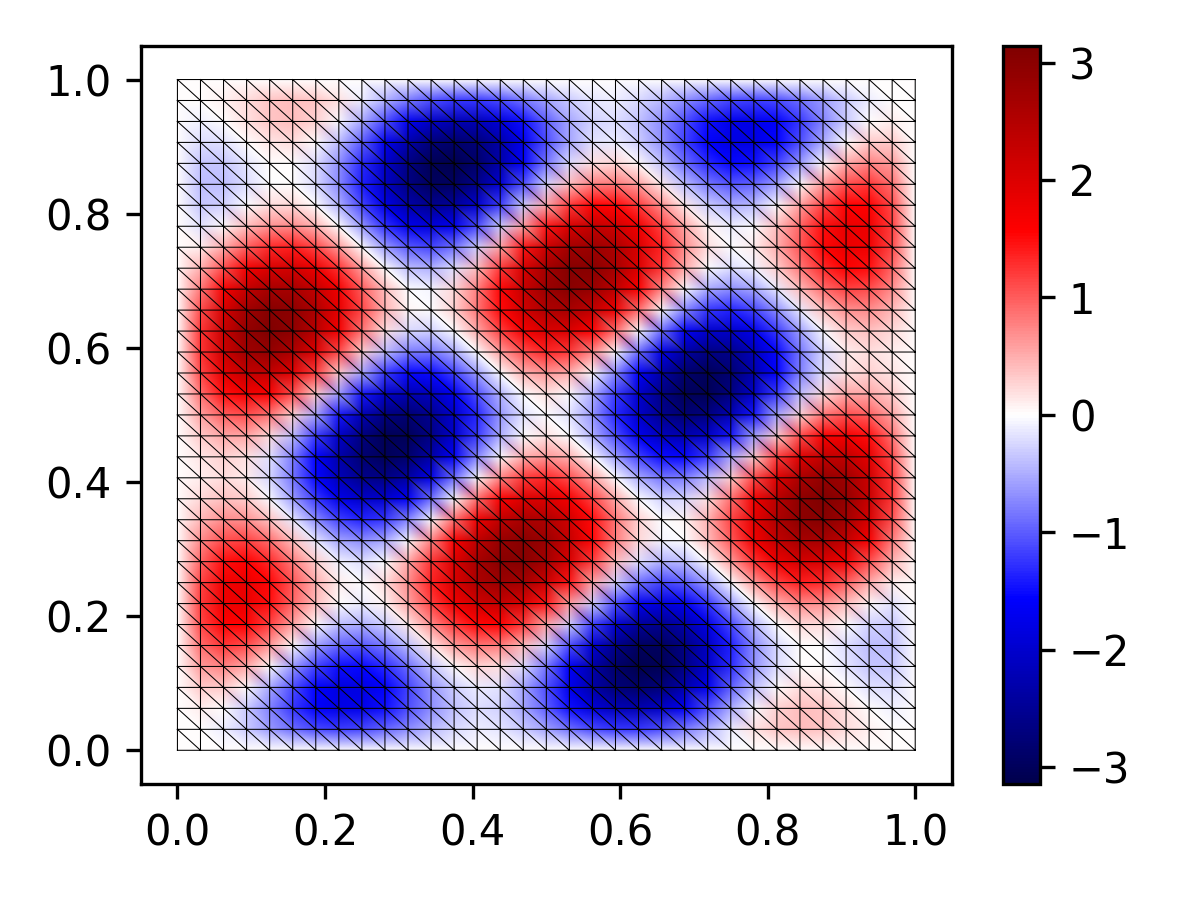}}
\caption{Finite element mesh on unit square domain $\Omega$, and the solution to the example PDE.}
\label{example_problem_fig}
\end{figure}

Here, we very briefly review the spiking finite element algorithm described more completely in \cite{theilman2025solving}. Given a linear, elliptic PDE defined on some 2- or 3-D domain with prescribed boundary conditions, we discretize the domain with a regular or irregular mesh. On this mesh, we construct a finite-dimensional function space by defining basis functions supported on the mesh's elements. Applying the Galerkin approximation to the weak form of the PDE, we arrive at a large, sparse linear system for the coefficients of the best approximation to the PDE solution within the finite-dimensional approximation space supported by the mesh. Thus, a PDE such as the Poisson equation
\begin{equation}
    \nabla^2 u = f
\end{equation}
becomes

\begin{equation}
    Ax = b
\end{equation}
Where $A$ is an $N \times N$ matrix where $N$ is the number of degrees of freedom (related to the number of elements in the mesh).
Importantly, because of the locality of the interactions in the physics described by the PDE, the matrix A is highly sparse, and the nonzero elements correspond to interactions between neighboring degrees of freedom in the mesh. 

Our neuromorphic finite element algorithm (NeuroFEM) embeds this sparse linear system into a network of spiking neurons. Each mesh node is represented by a collection of neurons that project to "readout" variables with either positive or negative sign and magnitude $|\Gamma|$. The projection from neurons to readout variables is represented by the $N$-variable by $M$-neuron matrix $\Gamma$. We associate $\textsc{NPM}$ neurons per mesh node, so $M = N \times \textsc{NPM}$.  The readout evolves in time according to the first-order dynamical system
\begin{equation}
    \frac{dx}{dt} = -\lambda_d x + \Gamma s(t)
\end{equation}
Where $s(t)$ is a vector of spikes from each neuron. 

The weights between neurons are proportional to the product $\Gamma^T A \Gamma$, which projects the $N \times N$ matrix $A$ to an $M \times M$ matrix of weights between neurons. Importantly, this implies that the connections between neurons share the same sparsity and locality present in the system matrix $A$. 

The individual neurons operate with generalized leaky-integrate-and-fire dynamics such that the readout variables driven by spikes approximate the dynamical system
\begin{equation}
    \frac{dx}{dt} = b - Ax
\end{equation}
Thus, assuming the eigenvalues of $A$ are all the same sign, the network's dynamics flow to the fixed point of this equation, precisely when $Ax = b$ and the linear system is solved. 

At their core, the neurons in NeuroFEM act as feedback proportional--integral (PI) controllers \cite{theilman2025solving}. Thus, like feedback controllers in other engineering contexts, each neuron is able to recalibrate its own firing in response to external perturbations in the whole circuit.

\subsection{FEM example problem}

In all of our experiments unless otherwise indicated, we construct a finite element approximation for the Poisson problem on a unit square domain $\Omega$ with Dirichlet boundary conditions, linear elements, and a fixed right-hand side. Specifically, we solve for $u$ in
\begin{align}
    \nabla^2 u(x, y) &= \sin(3\pi (x - y))\sin(2\pi(x+y)) &\text{on $\Omega$} \\
    u(x,y) &= 0 &\text{on $\partial \Omega$}
\end{align}

Figure \ref{example_problem_fig} illustrates the domain, mesh, and finite-element solution using a conventional solver. This example was chosen to ensure that large and small magnitude, positive and negative values were present in the solution across the mesh, to assess the impact of our perturbations on the full dynamic range of the circuit. 

\section{Ablating Neurons}

Our brains are resilient to neuron loss. Human brains lose approximately one neuron per second \cite{pakkenberg1997neocortical, pakkenberg2003aging}, and the brain shows enough plasticity to reorganize, preserve, or regain some meaningful level of function even after severe trauma. Recapitulating such resistance to injury is a highly desirable feature for brain-inspired neuromorphic algorithms. 

\begin{figure*}[htbp]
\centerline{\includegraphics[width=0.8\textwidth]{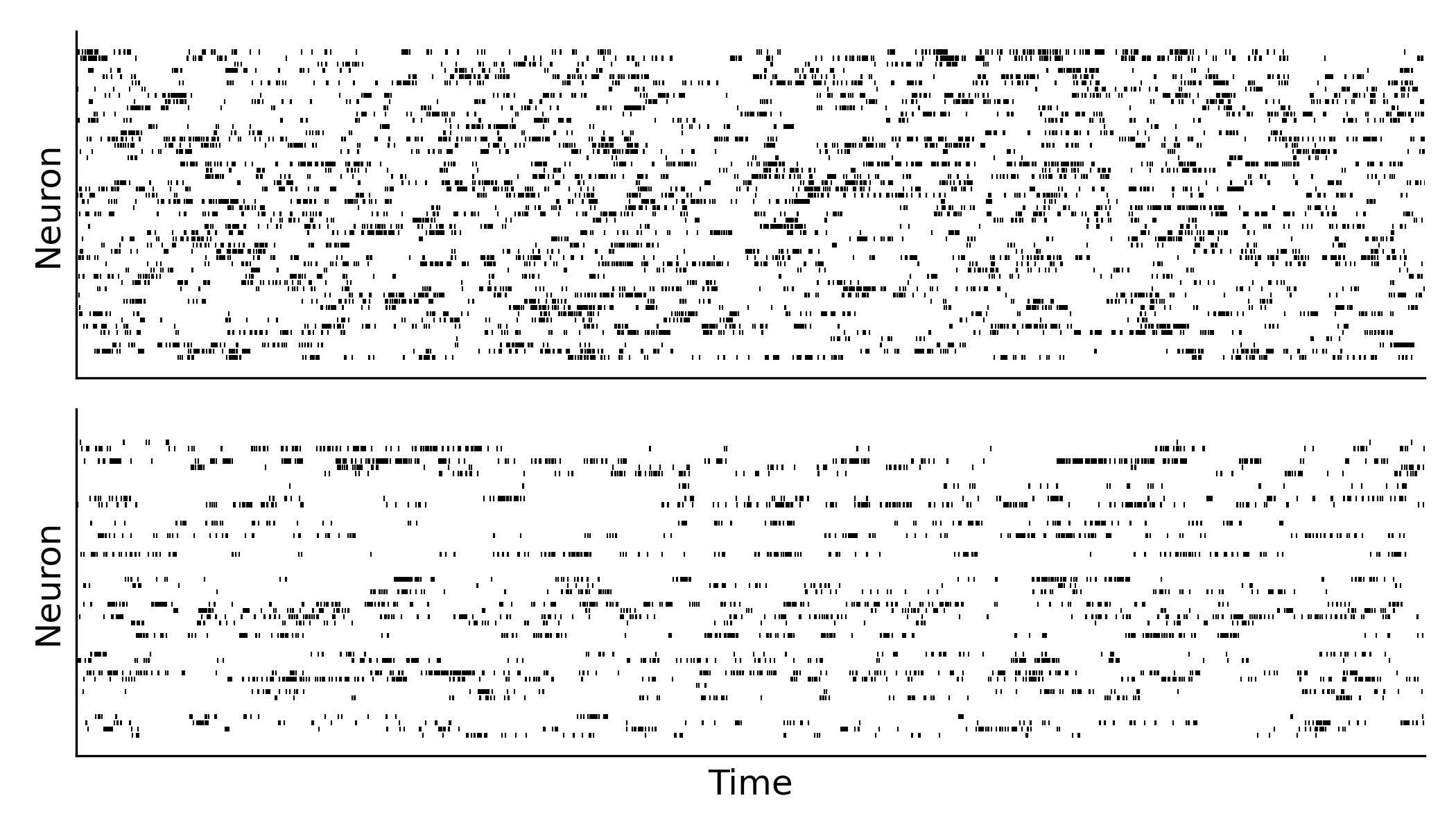}}
\caption{Ablating neurons. Top: Unmodified NeuroFEM raster plot solving the example problem. Shown are spike trains from 50 randomly chosen neurons across the mesh. Bottom: NeuroFEM raster plot with 40\% of the neurons ablated. }
\label{ablated_raster}
\end{figure*}

In our NeuroFEM algorithm, because individual variables are overrepresented with $\textsc{NPM}$ neurons per mech point, randomly ablating neurons in the network should result in little loss of accuracy until the remaining neurons per mesh node cannot adequately represent the solution. To test this, we ran trials in which we randomly ablated neurons before starting the network. For each trial, we generated a mask in which each neuron was independently ablated with probability $p$. Ablated neurons had their state variables fixed to zero and no spikes were emitted from these neurons during the simulation, as shown in Figure \ref{ablated_raster}  We then computed the relative error between the ablated network's solution and the solution produced by a conventional CPU solver on each trial. We averaged this relative error over trials to obtain an average performance for a specific ablation percentage. 

\begin{figure}[htbp]
\centerline{\includegraphics[width=0.5\textwidth]{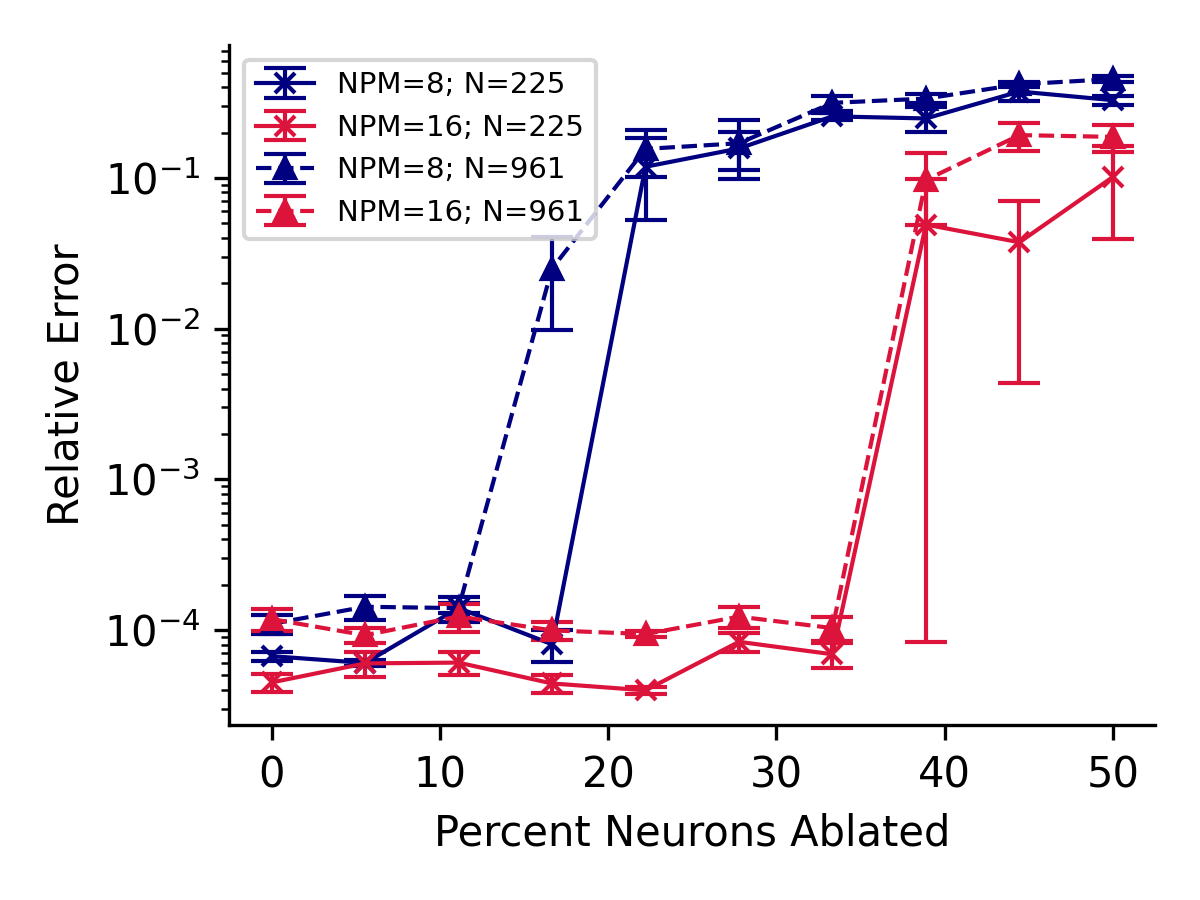}}
\caption{Relative error as a function of percent ablated neurons. Overall accuracy is unchanged until the proportion of ablated neurons reaches a threshold that depends on the number of neurons per mesh node (NPM). Increasing the redundancy by increasing the number of neurons per mesh node increases this threshold. Here, $N$ indicates the number of elements in the mesh and the size of the linear system. The total number of neurons in the circuit is $N\times \textsc{NPM}$. Error bars indicate standard error of the mean over 5 trials.}
\label{ablate_fig}
\end{figure}

As neurons are ablated, remaining neurons increase their firing rates to compensate. Eventually, this leads to individual neurons firing too fast with respect to the dynamical timescales of the circuit (i.e., to compensate would more than one spike per timestep), leading to a degradation in accuracy. In Figure \ref{ablate_fig}, we see that the NeuroFEM algorithm is robust to ablating neurons, and as the number of neurons per mesh node increases, the network becomes more robust. For 16 neurons per mesh node, as many as 32\% of the neurons may be ablated before a significant loss in accuracy results. The wide error bars show that it is also possible for a specific ablation pattern to result in little loss of accuracy even beyond this limit. While we only examined two values of neurons per mesh node, our results suggest that increasing this parameter would improve robustness even further. 

\begin{figure}[htbp]
\centerline{\includegraphics[width=0.5\textwidth]{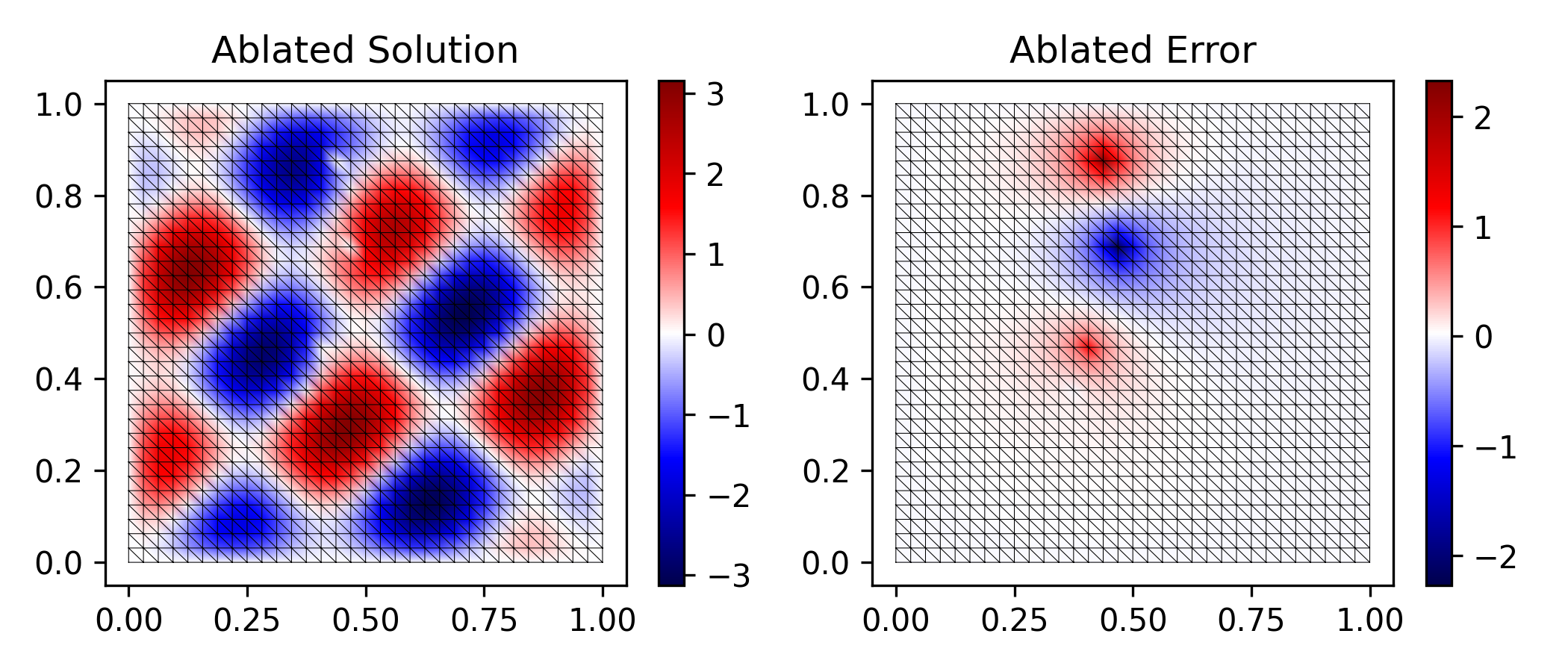}}
\caption{ Left: NeuroFEM solutions with 50\% neurons ablated. Despite clear errors at some mesh nodes, the overall solution remains close to the true solution. Right: Difference between the true and ablated NeuroFEM solutions. Error is concentrated to a few mesh nodes where, by chance, too many neurons were ablated.}
\label{ablate_error}
\end{figure}

Even though the relative error plot indicates significant error at 50\% ablated neurons, in Figure \ref{ablate_error} we see that this error is concentrated at only a few mesh nodes where, by chance, too many neurons were removed. Across the rest of the mesh, the ablated NeuroFEM solution remains close to the true solution; despite the large magnitude of the relative error, the solution degrades gradually. The ablated mesh nodes may be viewed as localized perturbations to the right-hand side of the example problem. Thus, the global effect of these ablated mesh nodes will depend on the mathematical properties of the PDE being solved. Here, because we are solving a Poisson equation, the error effectively diffuses over the mesh. 

\section{Dropping Spikes}

\begin{figure*}[htbp]
\centerline{\includegraphics[width=0.8\textwidth]{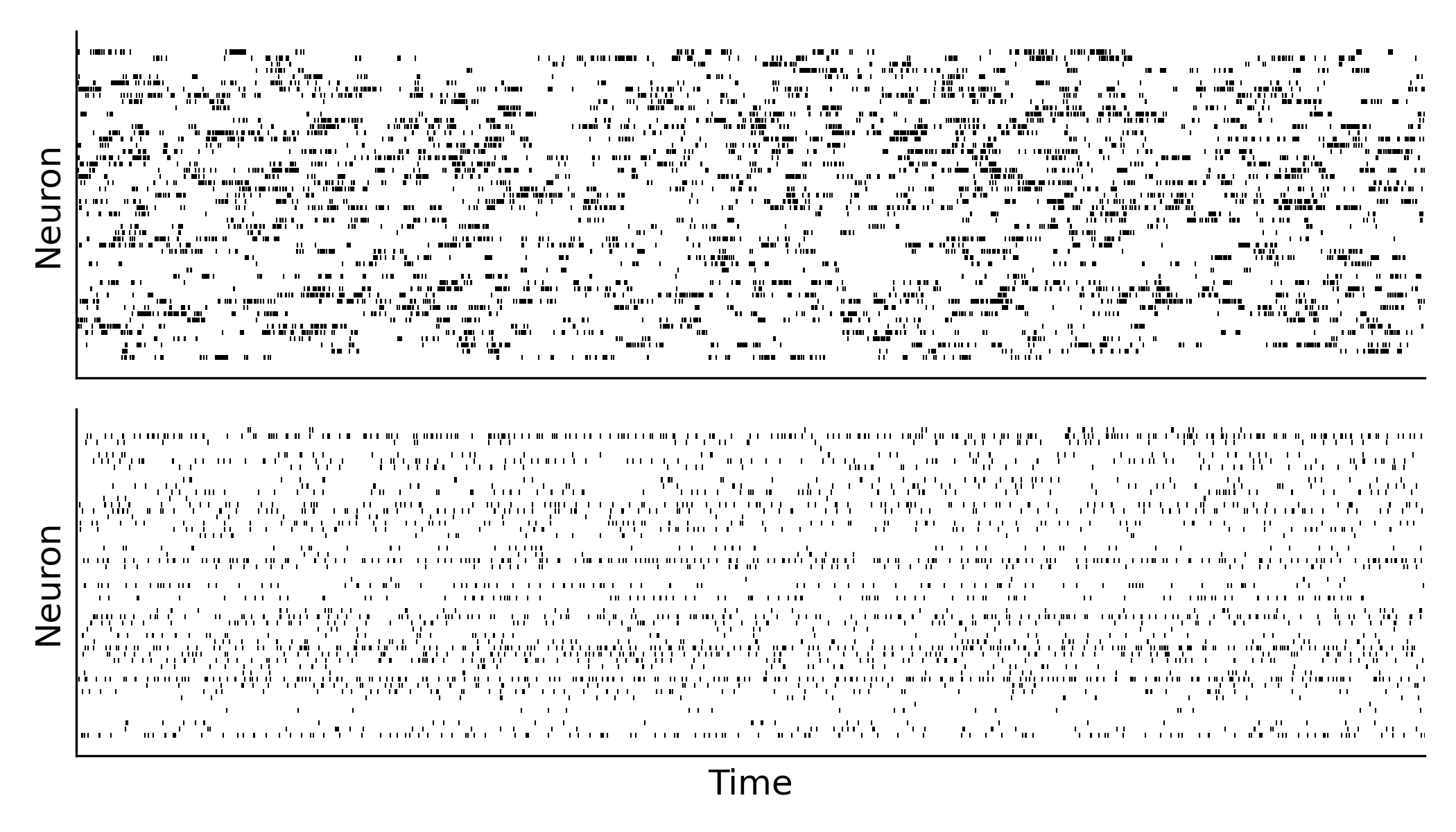}}
\caption{Dropping spikes. Top: Unmodified NeuroFEM raster plot solving the example problem. 50 randomly chosen neurons across the mesh are shown. Bottom: NeuroFEM raster plot with 90\% spikes dropped. Individual neurons recalibrate their firing to maintain solution accuracy. Dropping spikes acts like a regularizer leading to sparser activity, which may lead to a neuromorphic advantage.}
\label{dropped_spikes_raster}
\end{figure*}

Communicating spikes efficiently is a core design principle of neuromorphic hardware. Even so, it is possible that this communication may fail. Furthermore, stochastic spike transmission is a proposed neuromorphic hardware feature that is expected to yield advantages for certain algorithms \cite{aimone2022review}. We next examined how robust NeuroFEM is to stochastically dropped spikes. To do this, we used the same example problem as above, and repeated our experiment again for five trials. On each timestep, after generating spikes, we independently chose a random subset of neurons with probability $p$ and eliminated any spikes from those neurons. Thus, if a neuron decides to spike on any given timestep, that spike has a probability $p$ of not reaching any destination. Figure \ref{dropped_spikes_raster} shows spikes rasters from unmodified NeuroFEM and NeuroFEM with a high proportion of spikes dropped. 

\begin{figure}[htbp]
\centerline{\includegraphics[width=0.5\textwidth]{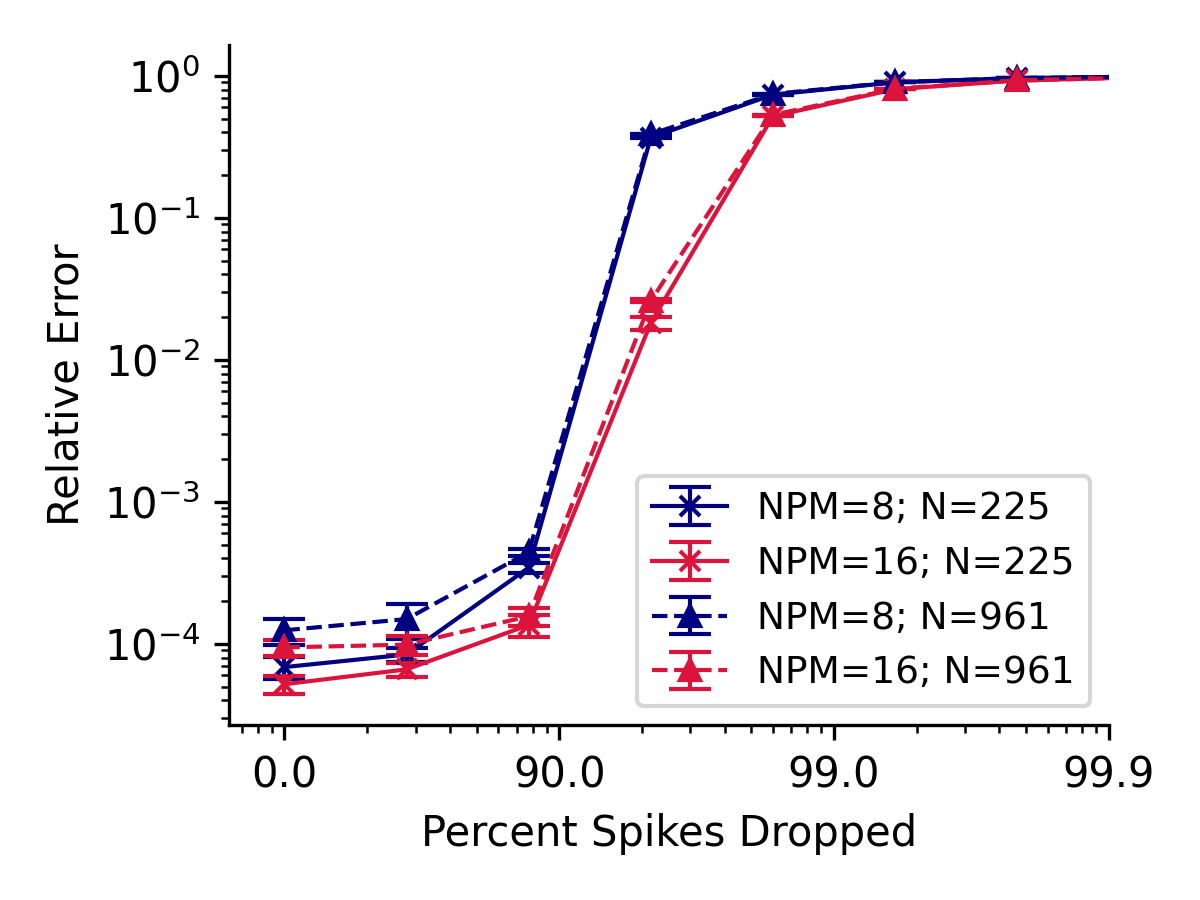}}
\caption{Relative error as a function of the proportion of spikes dropped on each timestep. Up to $\sim$90\% of spikes may be dropped on each timestep before the network shows an appreciable loss in accuracy for this example problem. Again, increasing the redundancy increases the robustness. Error bars indicate standard error of the mean over 5 trials.}
\label{drop_spikes_fig}
\end{figure}

We found that for $p$ as large as 90\% the NeuroFEM network shows no appreciable (i.e. within an order of magnitude) loss of accuracy (Figure \ref{drop_spikes_fig}). Based on the interpretation of the circuit as a network of PI controllers, as $p$ increases the individual neurons recalibrate their firing rates to offset the missing spikes. This continues until too many spikes per timestep are dropped and the network is unable to compensate because too little information is transmitted between neurons. 

\section{Discussion}
We have demonstrated intrinsic robustness and error tolerance in a natively neuromorphic algorithm for scientific computing. Our results show that it is precisely the brain-inspired elements of the algorithm that confer this tolerance, underscoring the brain as an important source of engineering inspiration. 

\subsection{Implications for a robust scientific computing}
Previous work looking at the robustness of iterative numerical solvers such as conjugate gradient on classical computers has noted that even single bit flips may derail the algorithm altogether \cite{elliott2014evaluating, agullo2020soft}. While it is possible to design algorithmic mitigations for these perturbations, these mitigations still rely on assumptions that the events are rare and specific \cite{elliott2014evaluating}. While this work does not definitively declare NeuroFEM as an uncontested winner in terms of algorithmic robustness, we have shown that the NeuroFEM algorithm has a different relation to computational faults than classical algorithms, with surprisingly wide tolerance bands. Thus, this work serves as a base case to start a longer conversation around robustness in neuromorphic algorithms and potential advantages compared to conventional computing. 

\subsection{Implications for a neuromorphic advantage}
Our results from stochastically dropping spikes hint at surprising possible neuromorphic advantages for the NeuroFEM algorithm. Because we found that up to 90\% of the spikes in the original algorithm may be dropped without consequence, it may be desirable to include this stochastic spike transmission directly into a hardware implementation of the algorithm. This would significantly reduce the required spike bandwidth, potentially leading to increased energy efficiency and reduced latency. Rather than viewing dropped spikes as a problem to be mitigated, it becomes an important feature. The probability of transmitting a spike when one is implied could be used as a ``throttle'' to dynamically tune the algorithm's performance, without destroying the core functionality of the algorithm. Future work will investigate this possibility on current digital CMOS neuromorphic platforms and on hypothesized analog implementations of the NeuroFEM circuit.

\subsection{Implications for spike representations}
A common question in neuromorphic computing is how to represent numerical values with spikes. The NeuroFEM algorithm uses a subtle but important strategy that is key to both its algorithmic effectiveness and our observed robustness. While it is true that at steady state, the number of spikes in a unit of time is proportional to the value of the represented solution value, NeuroFEM does \textit{not} use a ``rate code''. This is because the neurons within a mesh node coordinate between themselves using mutual inhibition to ensure that when one neuron spikes (by chance), its influence is subtracted from the other neurons. This means each neuron is not wholly and independently responsible for representing the readout variable; instead the responsibility is distributed across all the neurons. Previous work has pointed out that this kind of coordination is necessary to allow numerical values to be represented at high accuracy with reasonable numbers of spikes \cite{boahen2017neuromorphs}. Without this coordination, the number of spikes required to represent a value would grow quadratically with accuracy. However, for NeuroFEM it grows linearly \cite{theilman2025solving}. This coordination also allows the neurons to rebalance themselves due to the kinds of perturbations we applied in this study. 

Other kinds of spike-based encodings, such as time-to-first-spike, are also brittle to the kinds of perturbations examined here. Primarily, these encodings demand individual neurons carry the responsibility of a direct representation of the output value. In contrast, NeuroFEM neurons represent values implicitly:  the timing of individual spikes matters, but the ``meaning'' a spike with respect to the algorithm is only determined \textit{in relation to other spikes}. We contend that this is a fundamental concept necessary for any effective neuromorphic algorithm.

\subsection{Conclusions}
A final, perhaps encouraging point to conclude. We will note that NeuroFEM was not designed specifically for the robustness properties exhibited here. Instead, we found that design choices made, driven in large part by NeuroFEM's brain inspiration \cite{boerlin2013predictive}, in conjunction with the nature of the problem solved (finite elements), led to a spiking network with the requisite redundancy. Thus, not all neuromorphic algorithms will share these properties. For example, we would expect that another neuromorphic algorithm for solving PDEs with random walks \cite{smith2020solving, smith2022neuromorphic} will likely show robustness to some faults, but not necessarily the same faults considered here. Likewise, the robustness of spiking neuromorphic algorithms for artificial intelligence applications remains an open question, and perhaps algorithm robustness should be more of a consideration when evaluating potential neuromorphic strategies for artificial neural networks.

Finally, our study did not completely enumerate the space of possible faults and their impact on NeuroFEM, and we did not characterize the types of errors associated with any particular hardware platform. In particular, we anticipate that the NeuroFEM algorithm may benefit from use of analog neuromorphic processing for the neurons within each mesh node, so the impact of analog noise in both weights and activations would be an important perturbation not considered here. Because NeuroFEM allows for redundant synapses between neurons, it is likely that independent analog noise may ``average out'' at the global algorithmic level. 

To conclude, NeuroFEM provides a useful baseline for answering questions about robust neuromorphic algorithms, and we believe the principles demonstrated will be generalizable to new neuromorphic algorithms in the future.

\section*{Acknowledgments}
We thank the Advanced Simulation and Computing program at the U.S. Department of Energy for supporting this research.

This article has been authored by an employee of National Technology \& Engineering Solutions of Sandia, LLC under Contract No. DE-NA0003525 with the U.S. Department of Energy (DOE). The employee owns all right, title and interest in and to the article and is solely responsible for its contents. The United States Government retains and the publisher, by accepting the article for publication, acknowledges that the United States Government retains a non-exclusive, paid-up, irrevocable, world-wide license to publish or reproduce the published form of this article or allow others to do so, for United States Government purposes. The DOE will provide public access to these results of federally sponsored research in accordance with the DOE Public Access Plan \href{https://www.energy.gov/downloads/doe-public-access-plan}{https://www.energy.gov/downloads/doe-public-access-plan}. SAND2026-18243C

\input{neurofem_robustness.bbl}

\end{document}